\documentclass[sigconf, screen=false, review=false, anonymous=false, authordraft=False]{acmart}

\usepackage[yyyymmdd]{datetime}

\usepackage{tabularx}
\usepackage{dcolumn} %Aligning numbers by decimal points in table columns
\newcolumntype{d}[1]{D{.}{.}{#1}}

\usepackage{subcaption}

\usepackage{todonotes}
\let\xtodo\todo
\renewcommand{\todo}[1]{\xtodo[inline,color=green!50]{#1}}

\AtBeginDocument{%
  }

\setcopyright{acmcopyright}
\copyrightyear{2023}
\acmYear{2023}
\acmDOI{XXXXXXX.XXXXXXX}

\acmConference[CHI '23]{ACM CHI Conference on Human Factors in Computing Systems}{April 23--28,
  2023}{Hamburg, DE}
\acmBooktitle{CHI '23: Proceedings of the workshop \textit{Socially Assistive Robots as Decision Makers: 
Transparency, Motivations, and Intentions},
 April 28, 2023, Hamburg, DE}

\begin{document}

\title{Designing Dynamic Robot Characters to Improve Robot-Human Communications}


\settopmatter{authorsperrow=3}

\author{Carl Oechsner}
\email{c.oechsner@lmu.de}
\author{Daniel Ullrich}
\email{daniel.ullrich@ifi.lmu.de}
\affiliation{%
  \institution{LMU Munich}
  \streetaddress{Frauenlobstr. 7a}
  \city{Munich}
  \country{Germany}}

\renewcommand{\shortauthors}{Oechsner et al.}

\begin{abstract}
%\todo{What is the specific problem addressed?}
%\todo{What have you done?}
%\todo{What did you find out?}
%\todo{What are the implications on a larger scale?}
Socially Assistive Robots navigate highly sensible environments, which place high demands on safety and communication with users. The reasoning behind an SAR's actions must be transparent at any time to earn users' trust and acceptance. Although different communication modalities have been extensively studied, there is a lack of long-term studies investigating changes in users' communication needs over time. Considering two decades of research in Human-Robot Communication, we formulate the need to design dynamic robot personalities to unveil the full potential of SARs.

\end{abstract}


\begin{teaserfigure}
  \includegraphics[width=\textwidth]{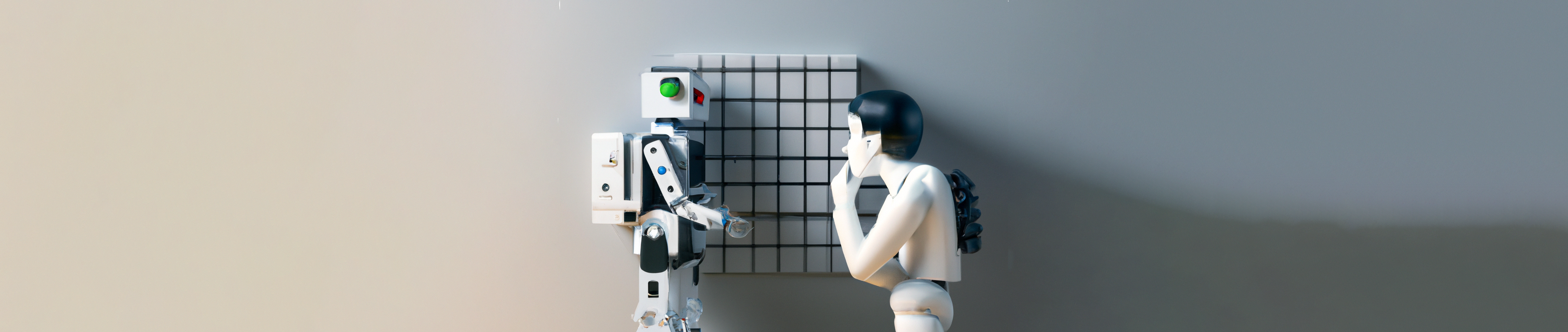}  
  \Description{A symbolic picture featuring a robot trying to explain itself to the user. Created with DALL-E using the prompt ``A photorealistic depiction of a person solving a puzzle together with a robot. both look at each other, trying to be polite, but obviously they do not understand each others actions'}
  \label{fig:teaser}
\end{teaserfigure}

%  \includegraphics[width=\linewidth]{sampleteaser}

\maketitle

\section{Introduction}

% First Paragraph
% CORE MESSAGE OF THIS PARAGRAPH:
%\todo{P1.1. What is the large scope of the problem?}
%\todo{P1.2. What is the specific problem?}

% Second Paragraph
% CORE MESSAGE OF THIS PARAGRAPH:
%\todo{P2. The second paragraph should be about what have others been doing}
%\todo{P2.3. Why is the problem important? Why was this work carried out?}

% Third Paragraph
% CORE MESSAGE OF THIS PARAGRAPH:
%\todo{P3.4. What have you done?}
%\todo{P3.5. What is new about your work?}

% Fourth paragraph
% CORE MESSAGE OF THIS PARAGRAPH:
%\todo{P4.6. What did you find out? What are the concrete results?}
%\todo{P4.7. What are the implications? What does this mean for the bigger picture?}

%Robots have made the leap from production lines to work alongside humans as cooperative robots in non-industrial contexts and are being used as social robots ...

While in 2005 \citet{feil-seifer_socially_2005} still defined that they assist merely through social interaction, in 2021 \citet{boada_ethical_2021} claimed that the application areas of Socially Assistive Robots (SARs) should also extend to robots that perform actions involving physical user contact.
% Working so close to humans and navigating their environments makes them need to navigate complex social interaction \cite{clabaugh_escaping_2019} and obey social rules \cite{aly_model_2013} to earn the users' acceptance and trust, which is crucial to use their potential.
Working in domains with vulnerable users, e.g., in-home care for the elderly, differentiated education for children, or mental health~\cite{clabaugh_escaping_2019}, requires them to navigate both complex environments, social interaction~\cite{clabaugh_escaping_2019}, and obey social rules~\cite{aly_model_2013} to earn the users' acceptance and trust. This is crucial to use the robot's full potential.

To make a system predictable, it has to have consistent patterns, behavior, and characteristics - or: a personality - that can be learned by the user to determine future behavior. Prior research has shown that a robot's perceived personality can affect trust and acceptance~\cite{salem_would_2015, walters_avoiding_2008}. In general, the tendency of people to attribute human traits to inanimate objects - anthropomorphism - affects robots too. Depending on the robot's behavior and appearance, users will attribute a ``mental'' internal state to it, which will in turn influence how they anticipate the robot to act~\cite{thellman_mental_2022}. That means by consciously designing a robot's personality, the designer can help users to understand the robot's reasoning and build up trust.

However, implementing studies with automatic robots is still complex and costly. Thus, researchers often have to fall back to Wizard-of-Oz techniques combined with teleoperating robots. This makes it difficult to study the long-term effects of human-robot interaction, as in this case the robotic systems do not work autonomously and have to be manually controlled by an operator~\cite{clabaugh_escaping_2019, breazeal_effects_2005}. Especially assistive robots, however, should be studied over longer periods of time to gain insights into the changing dynamics of human-robot relationships.

% is affected by communication, increasing trust, ....

% Since studies with robots are still complex and costly to implement, researchers draw on using programmed fixed routines, Wizard-of-Oz techniques, or teleoperating the robot. However, investigating long-term effects of an assistive robot that lives with the subject over weeks or months is impossible this way~\cite{clabaugh_escaping_2019, breazeal_effects_2005}. Existing long-term studies only give limited insights into the dynamics of human-robot relationships and how requirements change over time (c.f.~\cite {kidd_robots_2008}).

% \todo{relationship with robot}
% would need long-term studies, but...
% \begin{itemize}
%     \item SAR researchers overcome current technical barriers by fixed routines, WoZ, teleoperation BUT: unsuitable for long-term studies/interventions 
%     \item real-world environments are noisy
%     \item ML and robotics communities mostly don't work human-centered  \cite{clabaugh_escaping_2019}
%     \item 
% \end{itemize}

In sum, to clarify how trust and acceptance of a SAR can be achieved, we have to look at the combination of communication, personality, and relationship with the robot. %In chapter 3, we summarize and discuss how to approach robot behavior design holistically.

% Communication + Personality > Predictability > Trust > Acceptance

%SARs:
%\begin{itemize}
%\item Leap from production line/cages to work with/near humans, Cobots
    %\item Defined by \cite{feil-seifer_socially_2005}
    %\item subgroup of Socially Interactive Robots \cite{deng_embodiment_2019}
    %\item do not necessarily physically interact with their environment \cite{fong_survey_2002}
    %\item need for them to navigate complex social interaction \cite{clabaugh_escaping_2019} and obey social rules \cite{aly_model_2013}
    %\item example areas: in-home care, differentiated education, mental health \cite{clabaugh_escaping_2019}
    %\item "aims to supplement the efforts of clinicians, therapists, educators, and caregivers through individualized, socially mediated interventions with robots" \cite{clabaugh_escaping_2019}
    %\item SAR intervention is defined by desired outcome \& application domain
%\end{itemize}

\section{Related Work}
In the following, we will briefly summarize prior research on robot communication and transparency, robot embodiment, and types of robot-human relationships. 
\subsection{Reasoning and Communication}
For users to accept and trust robotic systems, they must be able to understand the reasoning behind the robot's actions. Therefore, the robot must be able to communicate its internal state and intentions to the user \cite{thellman_mental_2022}. Especially in collaborative tasks, non-verbal communication can remove the ambiguities of verbal exchange and increase task performance \cite{breazeal_effects_2005}. Furthermore, interactive social cues can help to achieve more social user responses \cite{ghazali_assessing_2019}, improve user experience \cite{terzioglu_designing_2020} and also help shape the perception of robot personality and emotion \cite{xu_hitchhikers_2023}. 

While the right choice of words, voice, pitch and volume are crucial for verbal interaction, other audible queues can be used to indicate and support the robot's reasoning \cite{read_situational_2014, zhang_nonverbal_2023, robinson_designing_2022}. In the following, we will briefly touch on further non-verbal communication modalities.

%\subparagraph{Example intentions:}
%Robots that move physically:
%\begin{itemize}
%    \item "I move because I want to go over there."
%    \item "I recognized this item and will move around it."
%    \item "I recognized the item I should get you."
%    \item "I wait because I cannot get around you."
%    \item "I want to give this object to you."
%\end{itemize}
% Robots that do not move physically:
% \begin{itemize}
%     \item "I recommend you take this object."
%     \item "I understood which object you meant."
% \end{itemize}

% \paragraph{Visual}
% \begin{itemize}
%     \item direct user's attention towards areas of interest, example: from diffuse light to accumulated light
%     \item color \cite{whittaker_designing_2021}
% \end{itemize}

\paragraph{Movement}
During collaboration, especially object handovers, humans communicate intent and timing mainly through posture and limb movement. Based on these observations, \citet{strabala_towards_2013} derived crucial elements for robot handovers. The robot should have a ``carrying posture'' that is highly distinguishable from other poses, so the willingness to hand an object is clearly recognizable even if the user is not currently focusing on the robot. In this pose, object and limbs are held close to the robot. To signal the handover intent, the robot should move the object towards the torso of the user, ideally holding it sideways and tilting it towards the user.

Even when the robot is inactive, the user needs to know when it is operable. When the robot is not moving, there is no telling apart from being switched off or inactive. \citet{breazeal_effects_2005} introduced an idle movement to their robot to signal ``aliveness'', and \citet{terzioglu_designing_2020} found that a ``breathing'' motion of their robotic arm is suitable to display its internal state and intent. In general, motions that are ``human-like'' are reported to have a positive notion and help users predict robot movements faster and more accurately compared to more direct or abstract movements \cite{hutchison_using_2013-1, strabala_towards_2013}.

% Robots that move in the same scope as the user \todo{necessary?}

\paragraph{Gestures}
Even robots with few movable extremities can achieve interpretable gestures (see R2-D2), like nodding or shaking for approval and refusal. Imitating the user's head movements can lead to more acceptance \cite{ghazali_assessing_2019} and a ``shrugging'' gesture can signal the user that an input could not be interpreted \cite{breazeal_effects_2005}. Gestures accompanying verbal output by the robot can determine the level of its perceived extraversion \cite{aly_model_2013} and therefore help shape its personality. While head gestures seem to have an engaging effect on users \cite{kose-bagci_effects_2009} and can convey emotional states like anger to the user \cite{ajibo_analysis_2020}, simply turning towards the user can signal attention \cite{whittaker_designing_2021}.

\paragraph{Gaze}
Gaze cues help communicate the robot's internal state and intent \cite{terzioglu_designing_2020}. In collaborative tasks, gaze can help establish grounding, disambiguation of spoken information, joint attention that signals understanding, and turn-taking \cite{mehlmann_exploring_2014}. \citet{moon_meet_2014} found that handovers are significantly faster if the robot gazed toward the anticipated handover location. One might think this only applies to robots with face-like or even just eye-like features (like, e.g., \cite{breazeal_effects_2005}). However, in their work, \citet{terzioglu_designing_2020} demonstrate how gaze and posture cues can be easily achieved, even with a non-humanoid robot. In their studies, they used a robotic arm with a two-finger end effector and achieved sufficient cues by attaching a pair of glasses on top of it while pointing the fingers at the object in question.

% \paragraph{Auditive}
% \cite{read_situational_2014, zhang_nonverbal_2023, robinson_designing_2022}

% \textit{Mimics}:
% \cite{bartneck_interacting_2003}
% Problem:
% What about non-anthropomorphic robots? In their studies, \citet{fiore_toward_2013} found that even non-humanoid robots are able to use social cues as meaningful signals to users.

\subsection{Personality and Embodiment}
According to \citet{deng_embodiment_2019}, the physical embodiment of robots ``includes the internal and external mechanical structures, embedded sensors, and motors that allow them to interact with the world around them''. Compared to virtual representations, embodied robots affect user performance and perception of an interaction \cite{wainer_role_2006}: it increases compliance \cite{bainbridge_effect_2008}, social engagement and enjoyment \cite{wainer_embodiment_2007, lee_are_2006, bainbridge_benefits_2011}, improves cognitive learning \cite{leyzberg_physical_2012} and motor skills \cite{fridin_embodied_2014}, and increases user engagement in social \cite{fong_survey_2002}, educational \cite{toh_review_2016} and clinical \cite{broadbent_acceptance_2009} context.

Designing a robotic assistant does not stop at the visual appearance, number and functionality of extremities, level of human-likeness, size, color, and shape. Considering the strong impact assistant embodiment has, profound thought has to go into the ``how'' of the robot's actions: How and when does the robot move? How fast should it move, and how close should it approach the human? Are the movements abstract or more human-like? How are movements linked to other communication channels?

\citet{deng_embodiment_2019} propose a process for designing robot embodiment that considers the desired context. They suggest starting from the task a robotic assistant is to fulfill. According to \citet{mcgrath_methodology_1995}, collaborative tasks can be classified by four task natures: Generate, Choose, Negotiate, and Execute. Based on the task, decide which relation (or role) the assistant should have to the user (see \autoref{sect:relation}). The assistant's role falls between abstract (metaphorical) and literal (or realistic). The levels of abstractedness, task nature, and the chosen role later influence the level of autonomy and intelligence the users expect from the assistant.

\subsection{Relation and Habituation}
\label{sect:relation}
An assistant's relation with the user falls between subordinate and superior \cite{deng_embodiment_2019, clabaugh_escaping_2019}. A \textit{subordinate} role can signal that the assistant wants to learn from or be instructed by the user and is the least complex to implement. It can encourage empathy \cite{short_understanding_2017} and self-efficacy \cite{bartneck_interacting_2003, fischer_levels_2012}. The \textit{peer} meets the user on equal footing. It can learn from and correct the users and successfully engage them in cognitive competition \cite{deng_embodiment_2019}. The role which is most difficult to implement is the \textit{superior} \cite{fong_survey_2002}. It can be used to increase user compliance and achieves higher reliability and competence \cite{kennedy_robot_2015} and therefore is suitable, e.g., for coaching purposes.

When the user first uses the system, there is, of course, a novelty effect. The user is yet to learn, understand and trust the robot. In this phase, transparency has to be high, and parameters, like, e.g. action speed, have to be low. After a while, when the user has built up trust in the system and gained knowledge about its capabilities, reasoning can be dialed down, and speeds can be increased. Nevertheless, these are not the only parameters that have to be adapted over time. A study by \citet{salter_robots_2004} has shown that the engaging functions of a robot can deteriorate over time, especially when used in the wild.
% Problem:
% \begin{itemize}
%     \item Lack of long-term studies to study habituation effect
%     \item Problem of inconsistent appearances in prior research

% \end{itemize}
\section{Creating Dynamic Robot Personalities}

We learned that adequate communication is necessary for a robotic system and that the robot's personality can shape the quality of communication. A well-defined robot personality can help users understand the robot's reasoning. 
Mimicking human behaviors helps engagement and trust but does not have to be exact \cite{whittaker_designing_2021}.
Even abstract behavioral cues are sufficient to distinguish between different robot personalities \cite{ehrenbrink_google_2017}.
What personality should a robot have? Studies indicate that the preferred personality amplifies the user's traits \cite{whittaker_designing_2021}. Extroverts, for example, using more vivid and more frequent gestures during a conversation, also accept robots approaching closer during interaction \cite{robert_personality_2018}. However, other studies have found participants to prefer a character opposite to theirs \cite{buisine_influence_2009}.
In their study \citet{ferrandez_vicente_creating_2015} confirm that a robot's personality design directly affects its perceived intelligence and, more importantly, social intelligence. Asserting social intelligence is crucial for users to believe the robot is capable of making reasonable decisions.

How should robot personalities be designed? \citet{whittaker_designing_2021} suggest using classic persona design \cite{cooper_inmates_1999}. Starting from a persona, designers can combine personality traits that make robot behavior more predictable.
In general, users seem to react better to extrovert robot personalities \cite{syrdal_looking_2007} and perceive it as more socially intelligent \cite{ferrandez_vicente_creating_2015}. However, as mentioned earlier, most robot studies are short-term and are conducted in controlled environments. Given a short time frame, an extrovert character leaves a better and more memorable impression than an introvert one could. We assume that a robot companion with exclusively extravert behavior would be draining over a more extended period.

This is where the dynamic aspect of robot personality comes into play: 
We propose a more open personality emphasizing invitation and transparency for the first interaction phase (``getting-to-know''). 
After that, facets of extraversion and communication frequency, as well as an excessive amount of gestures, should be toned down, as well as explanations that serve transparency (e.g., explaining each time for repetitious tasks why the SAR reaches a particular stance in a decision-making process).
The result should be a smooth transition from the novelty phase, shaped by amazement considering the unknown functionalities, to the phase of habituation, in which the novelty effect is depleted, and users value a robust, reliable system.

In our view, one big challenge that this approach of dynamic personality poses is again rooted in anthropomorphism: We, as humans, value consistent personalities in other humans and are deterred by personality fluctuations. Changes in behavior or personality traits can hint at impostors - a link we do not want in the context of trust-building. Therefore, personality changes must be fine-tuned to fly under the radar - otherwise, we would change one drawback for another.

% -- Story "Transparenz nach Entscheidung": Wie kann das System adäquat kommunizieren, welche Entscheidung getroffen wurde und was gemacht wird.
% "Adäquat" bedeutet hier einerseits transparent (man will wissen, was passiert) und andererseits niederschwellig (man will nicht genervt sein).
% Hier könnte man fast die Arbeit von Laura mit der Aufmerksamkeitslenkung nehmen als Beispiel-Anwendung. 

% -- Story "Entscheidungen und Persönlichkeit": Das System muss einen bestimmten Systemcharakter besitzen (eine Rolle, die sich in Form eines Charakters ausdrückt), damit man dem System zugesteht, Entscheidungen treffen zu können: 

% Ein reiner Diener soll keine Entscheidungen treffen, er soll nur Befehle ausführen.
% Ein Companion darf Entscheidungen treffen (in bestimmtem Rahmen).
% Diener und Companion unterscheiden sich in Charakter und Auftreten. 

% -- Story "Trust-Building \& SAR": Neue Technologien haben unbekannte Eigenschaften (Überraschung! . Um Trust aufzubauen, muss das System in der Anfangsphase besonders transparent und vorsichtig agieren.
% Später - wenn ein Mindestmaß an Trust existiert - kann die Transparenz zurückgefahren werden und die Geschwindigkeit erhöht werden.
% In Learning-Sciences gibt es ein Konzept, dass sich Scaffolding nennt und was wir als Analogie missbrauchen können: Wir bauen ein Gerüst, das beim Lernen/Gewöhnen an das System hilft. Sobald eine Gewöhnung stattgefunden hat, wird das Gerüst nach und nach zurückgebaut.

\section{Conclusion}
We argue that during the design of SARs these vital must be taken into account to achieve transparent and trustful SARs: a coherent robot personality, that reflects in coherent behavior, movement, verbal and non-verbal communication as well as changing factors in human-robot relationship dynamics. More long-term studies have to be conducted that focus on the change of requirements to derive best practices on how the user can implicitly or explicitly control the amount of reasoning by the robot which, given the rapid development in AI techniques over the last years, is now more likely to happen than ever.
\begin{acks}
This project is funded by the Deutsche Forschungsgemeinschaft (DFG, German Research Foundation) – 425412993 and is part of Priority Program SPP2199 Scalable Interaction Paradigms for Pervasive Computing Environments.
\end{acks}

\bibliographystyle{ACM-Reference-Format}
\bibliography{main}

%\appendix

\end{document}